\documentclass[10pt]{article}
\usepackage{aaai}
\newcommand{\ul}{\underline}
\input{psfig}

\title{COMPUTING DIALOGUE ACTS FROM FEATURES WITH TRANSFORMATION-BASED LEARNING}
\author{Ken Samuel, Sandra Carberry, and K. Vijay-Shanker\\
	Department of Computer and Information Sciences\\
	University of Delaware\\
	Newark, Delaware 19716 USA\\
	\{samuel,carberry,vijay\}@cis.udel.edu\\
	http://www.eecis.udel.edu/\~{ }\{samuel,carberry,vijay\}/}

\begin{document}

\maketitle

\bibliographystyle{aaai}

\begin{abstract}
\begin{quote}

To interpret natural language at the discourse level, it is very
useful to accurately recognize dialogue acts, such as SUGGEST, in
identifying speaker intentions. Our research explores the utility of a
machine learning method called Transformation-Based Learning (TBL) in
computing dialogue acts, because TBL has a number of advantages over
alternative approaches for this application. We have identified some
extensions to TBL that are necessary in order to address the
limitations of the original algorithm and the particular demands of
discourse processing. We use a Monte Carlo strategy to increase the
applicability of the TBL method, and we select features of utterances
that can be used as input to improve the performance of TBL. Our
system is currently being tested on the \scshape VerbMobil \normalfont
corpora of spoken dialogues, producing promising preliminary results.

\end{quote}
\end{abstract}

\section{Introduction}

In order to properly understand a natural language dialogue (and,
potentially, to participate in the dialogue), a computer system must
be sensitive to the speakers' intentions. We will use the term, {\em
dialogue act}, to mean: a concise abstraction of the intentional
function of a speaker. The example dialogue in Figure~\ref{ex-das}
illustrates utterances that are tagged with dialogue acts. Note that,
in many cases, the dialogue act cannot be directly inferred from a
literal interpretation of the utterance.

\begin{figure}[ht]
\centering
\begin{tabular}{|rlc|}
\hline
$\mathrm{A_{1}:}$ & I have some problems                    & INFORM \\
                  & with the homework.                      & \\
$\mathrm{A_{2}:}$ & Can I ask you a couple                  & REQUEST \\
                  & of questions?                           & \\
$\mathrm{B_{1}:}$ & I can't help you now.                   & REJECT \\
$\mathrm{B_{2}:}$ & Let's discuss it Friday...              & SUGGEST \\
$\mathrm{A_{3}:}$ & Okay.                                   & ACCEPT \\
\hline
\end{tabular}
\caption{Dialogue between speakers A and B}
\label{ex-das}
\end{figure}

In recent years, people have begun to investigate methods for
assigning dialogue acts to utterances. Many of these researchers, such
as Hinkelman~\cite{Hinkelman90}, have followed traditional natural
language processing paradigms, analyzing corpora of dialogues by hand
and using intuition to derive general principles for recognizing
intentions. Two problems arise with this approach: Analyzing enough
data to uncover the underlying patterns may take too much time, and it
may be very difficult to recognize all of the relevant features and
how they interact to convey dialogue acts. As a result, these sets of
rules are likely to have errors and omissions.

Recently, a new paradigm has been emerging, in which machine learning
methods are utilized to compute dialogue acts. Machine learning offers
promise as a means of associating features of utterances with
particular dialogue acts, since the computer can efficiently analyze
large quantities of data and consider many different feature
interactions. A number of machine learning techniques have been
applied to this problem, but they have had limited success. One
possible explanation is that these approaches don't take full
advantage of particular features of utterances that may provide
valuable clues to indicate the dialogue acts.

This paper will begin with a survey of the other projects that have
used machine learning methods to compute dialogue acts. Then, we will
describe a relatively new machine learning algorithm called
Transformation-Based Learning (TBL), and investigate its merits for
the task of recognizing dialogue acts. We will also identify a few
limitations of TBL and address them with a set of features to help
distinguish dialogue acts and a Monte Carlo strategy to improve the
efficiency of TBL. We will then report some promising preliminary
experimental results from the system that we have developed, and we
will outline our plans for future improvements. Finally, we will
conclude with a discussion of this work.

\section{Current Approaches}

A number of researchers have reported experimental results for machine
learning algorithms designed to compute dialogue acts.
Figure~\ref{rel-res-summary} summarizes these experiments using the
following parameters:

\begin{figure*}
\centering
\begin{small}
\begin{tabular}{c|c|c|c|c|r|r|r|l|c}
Task & Features & Languages & Algorithm & Med. & \multicolumn{1}{c|}{Tags} &
\multicolumn{1}{c|}{Train} & \multicolumn{1}{c|}{Test} &
\multicolumn{1}{c|}{Success} & Citation\\
\hline
\hline
pred. & tags & Ger./Eng. & tag NGs & ftf & 42 & 105 & 45 & 30\% & Reithinger 1996\nocite{Reithinger96} \\
pred. & tags & Jap./Eng. & tag NGs & key. & 15 & 90 & 10 & 39.7\% & Nagata 1994a\nocite{Nagata94a} \\
pred. & tags & Ger./Eng. & tag NGs & ftf & 18 & 105 & 45 & 40\% & Reithinger 1996\nocite{Reithinger96} \\
pred. & tags & Ger./Eng. & tag NGs & ftf & 17 & 41 & 200 & 40.28\% & Reithinger 1995\nocite{Reithinger95} \\
pred. & tags & Ger./Eng. & tag NGs & ftf & 17 & 52 &  41 & 40.65\% & Alexandersson 1995\nocite{Alexandersson95a} \\
pred. & tags & Ger./Eng. & tag NGs & ftf & 17 & 52 & 81 & 44.24\% & Alexandersson 1995\nocite{Alexandersson95a} \\
pred. & tags & Jap./Eng. & tag NGs & tel. & 9 & 50 & 50 & 61.7\% & Nagata 1994b\nocite{Nagata94b} \\
\hline
comp. & tags/words/length & German & SCTs & ftf & 17 & 171 & 43 & 46\% & Mast 1995\nocite{Mast95} \\
comp. & tags/words/length & English & SCTs & ftf & 17 & 45 & 11 & 59\% & Mast 1995\nocite{Mast95} \\
comp. & tags/words/speaker & German & word NGs & ftf & 43 & 350 & 87 & 65.18\% & Reithinger 1997\nocite{Reithinger97} \\
comp. & tags/words/speaker & German & word NGs & ftf & 18 & 350 & 87 & 67.18\% & Reithinger 1997\nocite{Reithinger97} \\
comp. & tags/words & English & word NGs & ftf & 17 & 45 & 11 & 67.3\% & Mast 1995\nocite{Mast95} \\
comp. & tags/words & German & word NGs & ftf & 17 & 171 & 43 & 68.7\% & Mast 1995\nocite{Mast95} \\
comp. & tags/words/speaker & English & word NGs & ftf & 18 & 143 & 20 & 74.7\% & Reithinger 1997\nocite{Reithinger97} \\
\end{tabular}
\end{small}
\caption{Systems that compute dialogue acts with
machine learning methods}
\label{rel-res-summary}
\end{figure*}

\begin{description}

\item[Task:] Some systems were developed to {\em predict} the next
utterance's dialogue act, in order to help interpret the utterance
when it arrives. Others {\em compute} an utterance's dialogue act
after the input has already been analyzed by the lower-level language
processes.

\item[Features:] When computing a given utterance's dialogue act, the
input to each of the systems included the dialogue act {\em tags} from
the preceding utterances. In addition, some systems utilized basic
features of the current utterance: specific {\em words} found in the
utterance, the utterance's {\em length} (number of words), and the
{\em speaker} direction (who is talking to whom).

\item[Languages:] These projects dealt with dialogues in German,
English, and/or Japanese.

\item[Machine Learning Algorithm:] Two different machine learning
algorithms have been implemented: 1) Semantic Classification Trees
(SCTs) and 2) N-Grams\footnote{In some cases, the system counted
n-grams of {\em tags} (dialogue acts), while other systems focused on
n-grams of {\em words} found within utterances.} (NGs), smoothed with
deleted interpolation.

\item[Medium of Communication:] The dialogues took place face-to-face,
across a telephone line, or from keyboard to keyboard through a
computer network.

\item[Number of Tags:] This column specifies how many different
dialogue acts were used to label the corpora under analysis.

\item[Training and Testing Set Sizes:] These values represent the
number of dialogues in the tagged corpora that were used for training
and testing the systems.

\item[Success Rate:] After training, each system attempted to label the
data in the testing set. These numbers represent the best
reported scores.

\item[Citation:] The final column provides pointers to the appropriate
papers.

\end{description}

Based on these results, it appears that the most significant factors
are the task of the system and the features used, followed by the type
of machine learning algorithm and the number of different tags under
consideration. In this paper, we will present a system that uses TBL
to compute dialogue acts with several features of utterances that
these previous approaches did not consider.

\section{Transformation-Based Learning}

Brill introduced the TBL method and showed that it is very effective
on the part-of-speech tagging problem\footnote{This syntactic task
involves labeling words with part-of-speech tags, such as Noun and
Verb.}; it achieved accuracy rates as high as 97.2\%, which is as good
as or better than any other results reported for this
task~\cite{Brill95a}. Computing part-of-speech tags and computing
dialogue acts are similar processes, in that a part-of-speech tag is
dependent on the surrounding words, while a dialogue act is dependent
on the surrounding utterances. For this reason, we believe that TBL
has potential for success on the problem of computing dialogue acts.

\subsection{Labeling Data with Rules}

Given a training corpus, in which each entry is already labeled with
the correct tag, TBL produces a sequence of rules that serve as a
model of the training data. These rules can then be applied, in order,
to label untagged data.

\begin{figure}[ht]
\centerline{\psfig{figure=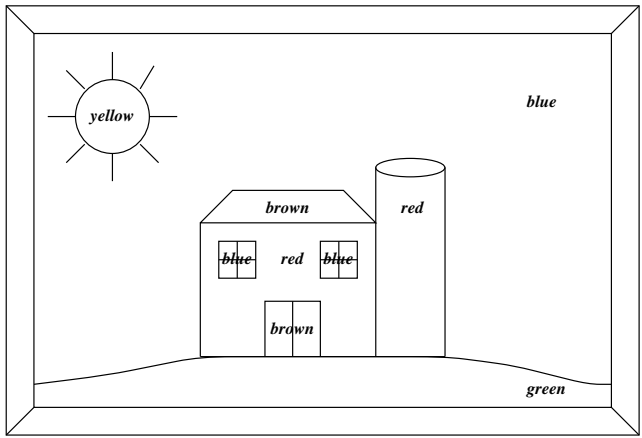}}
\caption{A barnyard scene}
\label{tbl}
\end{figure}

The intuition behind the TBL method can best be conveyed by means of a
picture-painting analogy.\footnote{We thank Terry Harvey for
suggesting this analogy.} Suppose that an artist uses the following
method to paint a simple barnyard scene. (See Figure~\ref{tbl}.) He
chooses to begin with the blue paint, since that is the color of the
sky, which covers a majority of the painting. He takes a large brush,
and simply paints the entire canvas blue. Then, after waiting for the
paint to dry, he decides to add a red barn. In painting the barn, he
doesn't need to be careful about avoiding the doors, roof, and
windows, as he will fix these regions in due time. Then, with the
brown paint, he uses a smaller, thinner brush, to paint the doors and
roof of the barn more precisely. Next, he paints green grass and a
yellow sun. He then returns to the blue to repaint the barn's windows.
And, finally, he takes a very thin, accurate brush, dips it in the
black paint, and draws in all of the lines.

The important thing to notice about this painting strategy is the way
that the artist begins with a very large, thick brush, which covers a
majority of the canvas, but also applies paint to many areas where it
doesn't belong. Then, he progresses to the very thin and precise
brushes, which don't put much paint on the picture, but don't make any
mistakes. TBL works in much the same way. The method generates a
sequence of rules to use in tagging data. The first rules in the
sequence are very general, making sweeping generalizations across the
data, and usually making several errors. Subsequently, more precise
rules are applied to fine-tune the results, correcting the errors, one
by one.

Figure~\ref{ex-rules} presents a sequence of rules that might be
produced by TBL for the task of computing dialogue acts. Suppose these
rules are applied to the dialogue in Figure~\ref{ex-das}. The first
rule is extremely general, labeling every utterance with the dialogue
act, SUGGEST. This correctly tags the fourth utterance in the sample
dialogue, but the labels assigned to the other utterances are not
right yet. Next, the second rule says that, whenever a change of
speaker occurs (meaning that the speaker of an utterance is different
from the speaker of the preceding utterance), the REJECT tag should be
applied. This rule labels utterances A$_{1}$\footnote{A change of speaker
always occurs for the first utterance of a dialogue.}, B$_{1}$, and A$_{3}$
with REJECT. The third rule tags an utterance INFORM if it contains
the word, ``I'', which holds for utterances A$_{1}$, A$_{2}$, and B$_{1}$. Next,
the fourth rule changes the tag on utterance A$_{2}$ to REQUEST, because
it includes the word, ``Can''.

\begin{figure}[ht]
\centering
\begin{tabular}{|c|l|c|}
\hline
\# & \multicolumn{1}{c|}{Condition(s)} & New Dialogue Act \\
\hline
\hline
1 & {\em none}             & SUGGEST \\
\hline
2 & change of speaker      & REJECT \\
\hline
3 & Includes ``I''         & INFORM \\
\hline
4 & Includes ``Can''       & REQUEST \\
\hline
5 & Prev. Tag = REQUEST    & REJECT \\
  & Includes ``can't''     &        \\
\hline
6 & Current Tag = REJECT   & ACCEPT \\
  & Includes ``Okay''      &  \\
\hline
\end{tabular}
\caption{A sequence of rules}
\label{ex-rules}
\end{figure}

At this point, only utterances B$_{1}$ and A$_{3}$ are incorrectly tagged.
As we continue, the rules get more specific. The fifth rule states that,
if the previous tag (the tag on the utterance immediately preceding
the utterance under analysis) is REQUEST, and the current utterance
contains the word, ``can't'', then the tag of the current utterance
should be changed to REJECT. In the sample dialogue, this rule applies
to utterance B$_{1}$. And finally, the last rule changes the tag on
utterance A$_{3}$ to ACCEPT, so that all of the tags are correct.

\subsection{Producing the Rules}

The training phase of TBL, in which the system learns the rules,
proceeds in the following manner:

\begin{ttfamily}

\begin{enumerate}

\item Label each utterance with an initial tag.

\item Until the stopping criterion is satisfied,\footnote{Typically,
the stopping criterion is to terminate training when no rule can be
found that improves the tagging accuracy on the training corpus by
more than some predetermined threshold~\cite{Brill95a}.}

\begin{enumerate}

\item[a.] For each utterance that is currently tagged incorrectly,
\begin{enumerate}

\item Generate all rules that correct the tag.
\end{enumerate}

\item[b.] Compute a score for each rule generated.\footnote{The score
measures the amount of improvement in the tagging accuracy of the
training corpus that would result from including a given rule in the
final model~\cite{Brill95a}.}

\item[c.] Output the highest scoring rule.

\item[d.] Apply this rule to the entire corpus.

\end{enumerate}

\end{enumerate}

\end{ttfamily}

This algorithm produces a sequence of rules, which are meant to be
applied in the order that they were generated. Naturally, some
restrictions must be imposed on the way in which the system may
generate rules in step 2ai, for there are an infinite number of rules
that can fix the tag of a given utterance, most of which are
completely unrelated to the task at hand.\footnote{For example, the
following rule would correctly tag utterance B$_{2}$ in
Figure~\ref{ex-das}: {\em IF} the third letter in the second word of
the utterance is ``s'', {\em THEN} change the utterance's tag to
SUGGEST.} For this reason, the human developer must provide the system
with a set of {\em rule templates}, to restrict the range of rules
that may be considered. Five sample rule templates are illustrated in
Figure~\ref{ex-templates}; these templates are sufficiently general to
produce all of the rules in Figure~\ref{ex-rules}. For example, the
last template can be instantiated with \ul{X}=REQUEST,
\ul{w}=``can't'', and \ul{Y}=REJECT to produce rule 5.

\begin{figure}[ht]
\centering
\begin{tabular}{|rl|}
\hline
{\em IF}   & {\em no conditions} \\
{\em THEN} & change \ul{u}'s tag to \ul{Y} \\
\hline
{\em IF}   & \ul{u} contains \ul{w} \\
{\em THEN} & change \ul{u}'s tag to \ul{Y} \\
\hline
{\em IF}   & change of speaker for \ul{u} is \ul{B} \\
{\em THEN} & change \ul{u}'s tag to \ul{Y} \\
\hline
{\em IF}   & the tag on \ul{u} is \ul{X} \\
{\em AND}  & \ul{u} contains \ul{w} \\
{\em THEN} & change \ul{u}'s tag to \ul{Y} \\
\hline
{\em IF}   & the tag on the utterance preceding \ul{u} is \ul{X} \\
{\em AND}  & \ul{u} contains \ul{w} \\
{\em THEN} & change \ul{u}'s tag to \ul{Y} \\
\hline
\end{tabular}
\caption{A sample set of templates,
where \ul{u} is an utterance, \ul{w} is a word,
\ul{B} is a boolean value, and \ul{X} and \ul{Y} are dialogue acts}
\label{ex-templates}
\end{figure}

\subsection{Justifications for Choosing TBL}

Decision Trees (DTs) and Hidden Markov Models (HMMs) are two popular
machine learning methods. Ramshaw and Marcus~\cite{Ramshaw94} compared
TBL with DTs and reported two advantages of TBL:

\begin{description}

\item[Leveraged Learning:] In the middle of a training session, TBL
can use the tags that have already been computed to help in computing
other tags, while DTs cannot make use of this type of information.

\item[Overtraining:] DTs tend to experience initial improvement, but
then, as training proceeds, performance degrades on unseen
data.\footnote{Although it is possible to prune a DT to address the
overtraining problem, this requires additional tuning.} TBL is
generally resistant to this overtraining effect, since its
rules are selected based on the {\em entire} training corpus, while each
DT rule only takes a subset of the training instances into account.
Ramshaw and Marcus presented experimental evidence to support the fact
that TBL is resistant to overtraining.

\end{description}

Ramshaw and Marcus also revealed a significant deficiency of TBL with
respect to DTs:

\begin{description}

\item[Largely Independent Rules:] When training with DTs, all of the
decisions (except the first) that the system makes depend directly on
choices made previously. But each decision made in training a TBL
system is largely independent of the earlier choices. This means that,
in general, TBL has more freedom than DTs, and, as a result, TBL must
be provided with information, in the form of rule templates, to
restrict its freedom. Unfortunately, these rule templates can be quite
difficult to derive.

\end{description}

\begin{figure*}
\centering
\begin{tabular}{c|l}
Feature & \multicolumn{1}{c}{Sample Values} \\
\hline
\hline
cue phrases        & ``but'', ``and'', ``so'',
``anyway'', ``please'', ``by the way'', ``okay'', ``I'', ``Can'', ``can't'' \\
change of speaker  & true, false \\
tags               & INFORM, REQUEST, SUGGEST, ACCEPT, REJECT,
SUPPORT, ARGUE \\
short utterances   & ``Okay.'', ``Yes.'', ``No.'', ``Hello.'', ``Sorry.'',
``Well.'', ``Oh.'', ``Sounds good.'' \\
utterance length          & 1 word, 2 words, 3 words,  more than 3 words, more than 10 words \\
\hline
punctuation        & period, question mark, comma, exclamation mark,
semicolon, dash, nothing \\
surface speech acts       & direct, imperative, interrogative-yes-no, interrogative-wh,
suggestive, other \\
subject type              & ``I'', ``you'', ``he'' or ``she'' or
``it'', ``we'', ``they'', ``who'',
``this'' or ``that'', other \\
verb type                 & future tense, modal, ``be'', other \\
closest repeated word     & previous utterance, 2 utterances back, 3 utterances back, 4 utterances back, none \\
closest interrelated word & previous utterance, 2 utterances back, 3 utterances back, 4 utterances back, none \\
\end{tabular}
\caption{Features}
\label{features}
\end{figure*}

TBL also has a number of advantages over HMMs: 

\begin{description}

\item[Intuitive Model:] Unlike HMMs, which represent a learned model
as a huge matrix of probability values, TBL produces a relatively short
list of intuitive rules. This is a very attractive aspect of the TBL
algorithm, because a researcher can analyze these rules by hand in
order to understand what the system has learned. Any insights he gains
might allow him to alter the learning methodology to improve the
system's performance. Thus, while HMMs can produce a working model of
a set of data, TBL additionally offers insights into a {\em theory} to
explain the data. This is especially crucial in discourse, as no
complete theory exists yet.

\item[Discarding Irrelevant Input:] If an HMM is given access to
information that happens to be irrelevant to the task at
hand,\footnote{For example, the third letter of the second word of
each utterance is unlikely to be relevant to the task of computing
dialogue acts.} its performance suffers. This is because the
irrelevant information interferes with the important features of the
input in a fully-connected network. But, as Ramshaw and
Marcus~\cite{Ramshaw94} showed experimentally, TBL's success is
largely unaffected by irrelevant features in the input. This is
because rules that consider relevant features generally improve the
tags in the training corpus, while the effect of rules without any
relevant features is completely random. Thus, relevant rules tend to
be chosen for the final model, since they generally receive higher
scores, and the irrelevant rules are avoided, so they have no effect
when the model is used to label unseen data. This aspect of TBL makes
it an especially attractive choice for discourse processing, as
researchers still disagree on what the relevant features are for
computing dialogue acts.\footnote{Several researchers have proposed
different features, as we discussed in previous work~\cite{Samuel96}.
But these sets of features are likely to have errors and omissions.}
But if the system has access to a large set of features that {\em
might} be relevant, it can {\em learn} which ones are really relevant,
and ignore the rest. So the human developer only needs to provide the
system with an overly-general set of features, and allow the learning
method to select a subset.

\item[Distant Context:] The basic assumptions of HMMs prevent the
analysis of the focus shifts that frequently occur in dialogue, while
TBL can take distant context into account quite easily, by including
features that consider preceding utterances.

\item[Overtraining:] As stated above, TBL does not tend to experience
an overtraining effect. Given sufficient training time, the method may
learn rules that overfit to the training data, but these rules are
necessarily very specific, and thus they have little or no effect on
the unseen data.\footnote{This is appropriate, since there is not much
evidence in these cases.} On the other hand, given too much training
time, HMMs overfit to the training data, and so they may have
difficulty generalizing to unseen data.

\end{description}

To summarize, TBL has several advantages in comparison with DTs and
HMMs, particularly on the task of computing dialogue
acts.\footnote{Although we realize that it would be beneficial to try
applying other machine learning algorithms to our data for direct
comparison, we have not yet had an opportunity to run these
experiments.} Thus, we have decided to try using TBL for this task.
But TBL also has a significant limitation: its dependence on rule
templates. This problem will be addressed in the next section.

\section{Using TBL to Compute Dialogue Acts}

TBL has not previously been applied to any discourse-level problems.
In lifting the algorithm to this new domain, it has been necessary to
revise and extend TBL to address the limitations of the original
algorithm and to deal with the particular demands of discourse
processing.

\subsection{Features}

Current approaches for computing dialogue acts with machine learning
methods have made little use of features. In some cases, the input is
presented in such an opaque format that the system cannot learn from a
tractable quantity of training data; in other cases, some relevant
information is not presented to the system at all.

Our approach is to select certain features that can easily be
extracted from utterances, and which we believe would allow our system
to learn dialogue acts more effectively. To pinpoint features that are
relevant for this task, researchers have traditionally analyzed data
by hand, using intuition. Figure~\ref{features} presents a subset of
the features suggested by several different
researchers~\cite{Samuel96,Hirschberg93,Lambert93,Chen95,Andernach96,Reithinger97,Mast95,Nagata94b,Alexandersson95a,Reithinger95,Reithinger96,Nagata94a}.\footnote{The
feature, tag, refers to the dialogue act that the system has chosen,
as opposed to the dialogue act that is known to be correct in the
training corpus.} We are currently examining the features listed in
the upper half of Figure~\ref{features}.

The use of features addresses a significant concern in machine
learning, called the sparse data problem.\footnote{The sparse data
problem says that no corpus is ever large enough to be an adequate
representation for all aspects of language.} This problem is
especially serious for discourse-level tasks, because the input
arrives in the form of full utterances, and there are an infinite
number of possible utterances. Since most utterances do not appear
more than once in a tractable quantity of data, it is impossible for a
machine learning algorithm to make appropriate generalizations from
data in this raw form. If relevant features of the utterances are
selected in advance, it should aid learning significantly.

In our experience, the cue phrases feature tends to be very effective.
Several researchers have previously observed that there are certain
short phrases, called {\em cue phrases}, that appear frequently in
dialogues and convey a significant amount of discourse information.
These researchers have each used traditional methods to produce a list
of cue phrases; a survey of these lists is presented in Hirschberg and
Litman~\cite{Hirschberg93}.

It may be possible to use the power of machine learning to generate an
effective list of cue phrases automatically. We are collecting cue
phrases by scanning the training corpus and counting how many times
each n-gram (n = 1, 2, or 3) of words co-occurs with each dialogue
act, selecting those n-grams with co-occurrence scores higher than a
predetermined threshold. We expect that, if an n-gram is frequently
associated with a dialogue act, it may be able to successfully predict
the dialogue act.

We find that this method of collecting cue phrases is very
general.\footnote{It generates cue phrases that have not been reported
in the literature, such as ``that sounds great''; phrases that aren't
cue phrases, but still have a lot of domain-specific predictive power,
such as ``make an appointment''; phrases that include extra
unnecessary words, such as ``okay that''; and phrases that aren't
useful for this task at all, such as ``the''.} But this is acceptable,
because we are primarily concerned about missing relevant
cue phrases, since errors of omission handicap the system. Errors of
commission are less of a concern, because TBL can learn to ignore
irrelevant information.

Additionally, we are experimenting with a method of clustering related
words together into semantic classes. For example, the system is
currently producing similar rules for the cue phrases: ``Monday'',
``Tuesday'', ``Wednesday'', ``Thursday'', and ``Friday''. If it knew
that these are all weekdays, it could capture the necessary patterns
in a single rule, and this rule would have five times as much training
data supporting it. Other semantic classes that may be effective
include: months, numbers, and ordinal numbers.

\subsection{A Monte Carlo Version of TBL}

In a task such as dialogue act tagging, it is very difficult to find
the set of all and only the relevant templates. We wish to relieve the
developer of part of this labor-intensive task. As mentioned above,
TBL is capable of learning which templates are relevant and ignoring
the rest, so the developer only needs to produce a general set of
templates that includes any features that might be relevant. However,
if TBL is given access to too many templates, it has too much freedom
to generate rules, and the algorithm quickly becomes intractable. The
problem is that, in each iteration, TBL must generate {\em all} rules
that correct at least one tag in the training corpus. Based on
experimental evidence, it appears that it is necessary to limit the
system to about 30 or fewer  templates. Otherwise, the memory and
time costs become so exorbitant that the training phase of the system
breaks down. While a deep linguistic analysis of the data might
identify the necessary templates, any errors of omission would have a
significant detrimental effect on the system. Thus, it is critical
that the  templates be chosen carefully.

We have been experimenting with a Monte Carlo strategy to relax the
restriction that TBL must perform an exhaustive search. In a given
iteration, for each utterance that is incorrectly tagged, only R of
the possible instantiations are randomly selected, where R is a
parameter that is set in advance. It should be clear that, as long as
R is relatively small, the efficiency of the algorithm is improved
significantly. Theoretically, if R is fixed, then increasing the
number of templates does not affect the training and memory
efficiency, since the number of rules being considered for each
iteration and each utterance is held constant. This claim has been
supported experimentally. Consequently, the system can train
efficiently with thousands of templates, as opposed to 30.

Our experiments show that, as long as R is sufficiently
large,\footnote{In our dialogue act tagging experiments, we found that
R=8 is sufficient for 4000 templates.} there doesn't appear to be a
significant degradation in performance. We believe that this is
because the best rules are effective on many utterances, so there are
many opportunities to find these rules. In other words, although the
random sampling will miss several rules, it is highly likely to
generate the best rules.

Thus, the Monte Carlo extension enhances TBL, so that it works
efficiently and effectively with thousands of templates, thereby
increasing the applicability of the TBL method. Further information
about this work is presented in another paper~\cite{Samuel98b}.

\section{Early Results \& Planned Improvements}

We have implemented the TBL algorithm outlined above, and we are
currently testing it on the \scshape VerbMobil \normalfont corpora of
face-to-face dialogues~\cite{Reithinger97}, which consist of dialogues
with utterances that have been hand-tagged with one of 42 dialogue
acts. We have been focusing on the corpus that Reithinger and Klesen
used in producing their best accuracy rate. (See the last row of
Figure~\ref{rel-res-summary}.) This corpus of English dialogues was
divided into two disjoint sets: a training set with 143 dialogues
(2701 utterances) and a testing set with 20 dialogues (328
utterances). We are clustering the 42 dialogue acts into a set of 18
abstract dialogue acts, as Reithinger and Klesen did in their
experiment.

Our TBL approach has produced success rates as high as 73.17\%. This
result is not statistically different from the highest score reported
Reithinger and Klesen ($\chi^2=0.20 \ll 3.84,\ \alpha=0.05$).
Currently, we are continuing to tune our system, and we have several
ideas to try to improve our results:

\begin{description}

\item[More Features:] We have pinpointed a number of features that
have not yet been implemented, which are listed in the lower half of
Figure~\ref{features}. We will investigate how these additional
features might improve our system's performance.

\item[Recycling Rules:] Each time the TBL algorithm is trained, it
begins with an empty set of rules and generates new rules from
scratch. But it may be useful to bootstrap the system, by initializing
the set of potential rules with rules that were selected for the final
model in the system's previous executions.

\item[Choosing Cue Phrases:] We are exploring other methods for
collecting cue phrases, including a strategy that aims to minimize the
entropy of the dialogue acts. Also, we have considered combining human
strengths and machine strengths, by letting the system choose a very
general set of cue phrases and then selectively removing cue phrases
from this set, by hand.

\end{description}

\begin{figure*}
\centering
\begin{tabular}{|cclc|}
\multicolumn{1}{c}{\#} & \multicolumn{1}{c}{Speaker} &
\multicolumn{1}{c}{Utterance} & \multicolumn{1}{c}{Dialogue Act} \\
\hline
1 & John & Delaware is playing basketball against Rutgers this weekend. & INFORM \\
2 & John & Shall we place a bet on Delaware?  & SUGGEST \\
3a & Mary & Well, Delaware is the home team...  & SUPPORT \\
\hline
1 & John & Delaware is playing basketball against Rutgers this weekend. & INFORM \\
2 & John & Shall we place a bet on Delaware?  & SUGGEST \\
3b & Mary & Well, Rutgers is the home team...  & ARGUE \\
\hline
\end{tabular}
\caption{Two dialogues that depend on world knowledge}
\label{world-knowledge}
\end{figure*}

\section{Augmentations of TBL}

We are considering several more revisions and extensions of
TBL to address the limitations of the original algorithm and the
particular demands of discourse processing.

\begin{description}

\item[Confidence Measures:] One limitation of TBL is that, unlike
Hidden Markov Models, it fails to offer any measure of confidence in
the tags that it produces. Such confidence measures are useful in a
wide variety of ways. For example, if the tags produced by the system
conflict with information coming from alternative sources, confidence
measures can be used to help resolve the conflict. We have proposed a
potential solution to this problem, which involves using the
Committee-Based Sampling method~\cite{Dagan95} in a novel way.
Essentially, the system is trained more than once, to produce a few
different but reasonable models for the training data. Then, given new
data, each model independently tags the input, and the responses are
compared. For a given tag, the confidence measure is a function of the
agreement among the different models on that tag.~\cite{Samuel96}

\item[Weakly-Supervised Learning:] When there is not enough tagged
training data available, we would like the system to be capable of
training with untagged data. Brill developed an unsupervised version
of TBL for part-of-speech tagging, but this algorithm requires
examples that can be tagged unambiguously, such as ``the'', which is
always a determiner~\cite{Brill95b}. Unfortunately, in discourse, we
have few unambiguous examples. But we intend to examine the potential
of the following weakly-supervised version of TBL. The system is first
trained on a small set of tagged data to produce a few models. Then,
given untagged data, it applies the models it has learned, to derive
dialogue acts with confidence measures. Those tags that are marked
with high confidence measures can be used as unambiguous examples to
drive the unsupervised version of TBL.

\item[Future Context:] In this research, we have primarily focused on
the task of understanding dialogue, but we would potentially like to
be able to modify our system for use in generation, so that the
computer can participate in dialogues. But then a new issue arises,
since the system currently requires access to a full dialogue in order
to properly tag utterances with dialogue acts. If it is to participate
in a conversation, then when it is the system's turn to speak, it must
be able to form a preliminary analysis of the incomplete dialogue. One
possible solution to this problem is to impose a constraint that
prevents the system from considering forward context in its rules.
Alternatively, the system could learn two sets of rules: rules to form
a preliminary analysis without the benefit of forward context, and
rules to refine the analysis, once the following utterances have been
heard.

\item[Incremental Mode:] Currently, the learning phase of the system
operates in batch mode, requiring all of the training data to be
presented at once. We would like to implement an incremental mode, so
that the system can refine its rules as more training data becomes
available.

\item[Tracking Focus:] Discourse is not completely linear, flowing
from one utterance to the next. Rather, focus shifts frequently occur
in dialogue. We believe that information about the discourse structure
can be used to help a machine learning algorithm compute dialogue
acts.

\end{description}

\section{Discussion}

We have explored the effectiveness of TBL in recognizing dialogue acts
and argued that TBL has a number of advantages over alternative
approaches, such as DTs and HMMs. A significant problem with TBL is
that the system must analyze a tremendous number of rules. We are able
to overcome this problem by utilizing a Monte Carlo strategy, whereby
TBL randomly samples from the space of possible rules, rather than
doing an exhaustive search. We have experimentally found that this
significantly improves efficiency without compromising accuracy.
Additionally, we consider the use of features to improve the input to
the system; other machine learning systems that compute dialogue acts
have made little use of features. Also, we are automatically
generating sets of cue phrases. Our preliminary results with the
\scshape VerbMobil \normalfont corpora are encouraging, particularly
in light of the fact that we have only recently begun to implement the
system, and we still plan to investigate several further improvements,
such as considering a more extensive set of features.

To date, no system has been able to compute dialogue acts with better
than 75\% accuracy. Certainly, we cannot hope to achieve 100\% success
on this problem until we find an effective way to encode the
common-sense {\em world knowledge} that is necessary in some cases. Two
examples are presented in Figure~\ref{world-knowledge}, where the
dialogue act of the last utterance in each dialogue cannot be
determined without knowing that the home team generally has an
advantage in a basketball game. But in spontaneous dialogues, we
believe that people {\em usually} incorporate various cues into their
utterances so that dialogue acts may be recognized without relying
heavily on this type of information. Thus, we expect that our system
can achieve high performance, despite the fact that it lacks world
knowledge. We also envision that our system may potentially be
integrated in a larger system with components that can account for
world knowledge.

\section{Acknowledgements}

We wish to thank the members of the \scshape VerbMobil \normalfont
research group at DFKI in Germany, particularly Norbert Reithinger,
Jan Alexandersson, and Elisabeth Maier, for providing us with the
opportunity to work with them and generously granting us access to the
\scshape VerbMobil \normalfont corpora to test our system.

This work was partially supported by the NSF Grant \#GER-9354869.

\bibliography{/usa/samuel/class/research/related_research/mybibfile}

\end{document}